\documentclass[12pt]{iopart}

\usepackage{iopams}  
\usepackage{graphicx}
\usepackage[frozencache,cachedir=.]{minted}
\usepackage{float}
\usepackage{subcaption}
\usepackage[english]{babel}
\usepackage[square,numbers]{natbib}
\usepackage{pifont}
\usepackage{xcolor}
\usepackage{moresize}
\usepackage{cprotect}

\definecolor{mygreen}{RGB}{63, 204, 101}
\newcommand{\cmark}{\color{mygreen}\ding{51}}%
\newcommand{\xmark}{\color{red}\ding{55}}%
\definecolor{bg}{rgb}{1.,1.,1.}

\begin{document}

\title[Slax: JAX Library for SNNs]{Slax: A Composable JAX Library for Rapid and Flexible Prototyping of Spiking Neural Networks}

\author{%
  Thomas M. Summe}
\address{Dept. of Computer Science and Engineering, University of Notre Dame}
\ead{tsumme@nd.edu}
\vspace{10pt}
\author{
  Siddharth Joshi}
\address{Dept. of Computer Science and Engineering, University of Notre Dame}
\ead{sjoshi2@nd.edu}

\begin{abstract}

Recent advances to algorithms for training spiking neural networks (SNNs) often leverage their unique dynamics. While backpropagation through time (BPTT) with surrogate gradients dominate the field, a rich landscape of alternatives can situate algorithms across various points in the performance, bio-plausibility, and complexity landscape. Evaluating and comparing algorithms is currently a cumbersome and error-prone process, requiring them to be repeatedly re-implemented. We introduce Slax, a JAX-based library designed to accelerate SNN algorithm design, compatible with the broader JAX and Flax ecosystem. Slax provides optimized implementations of diverse training algorithms, allowing direct performance comparison. Its toolkit includes methods to visualize and debug algorithms through loss landscapes, gradient similarities, and other metrics of model behavior during training.
\end{abstract}
\let\thefootnote\relax\footnote{Preliminary draft}

\section{Introduction}

Taking cues from the energy-efficient information processing of biological neural networks, Spiking Neural Networks (SNNs) hold the potential to improved energy-efficiency in implementing machine intelligence. This potential arises from the extremely sparse, spike-based communication and computation within SNNs. However, the same sparsity that promises energy-efficiency also makes them challenging to train using conventional deep learning frameworks. Indeed, SNN training is still an area of active research, where researchers must balance ease of expressibility (ease of experimentation) with performance (tight feedback loops). In a similar vein as early machine learning (ML) frameworks like Theano~\cite{theano}, Caffe~\cite{caffe}, and Torch~\cite{torch}, early SNN training packages like Auryn~\cite{auryn} and Brian~\cite{brian} offered several usability advantages~\cite{snn_comparison} which spurred SNN research. More recently developed SNN frameworks, including snnTorch~\cite{snnTorch}, Spiking Jelly~\cite{spikingjelly}, Spyx~\cite{spyx}, and Rockpool~\cite{rockpool}, have built on contemporary ML frameworks like~\cite{PyTorch, tensorflow, jax_paper, mlx}. These frameworks offer a diverse spectrum of capabilities balancing between \textit{ease of use},\textit{ ease of customization},\textit{ performance}, and \textit{scalability}.

Existing frameworks have primarily focused on enabling gradient-based learning through surrogate gradients~\cite{surrogates} with backpropagation through time~\cite{bptt}. This focus can unintentionally limit performance optimizations for alternative training algorithms. Research exploring and implementing diverse algorithms such as e-prop \cite{e-prop, e-prop_spinnaker} is limited by the `first-class citizen', which is crucial to advancing SNN research.

To address this gap and facilitate rapid exploration of diverse training algorithms, particularly ones prioritizing performance and efficiency, we have developed a new SNN framework on top of JAX and Flax. JAX, due to just-in-time compilation and NumPy-like interface offers an ideal platform for efficiently prototyping and implementing new learning rules. Slax provides not only the basic building blocks for SNNs but also a suite of tools for online training, algorithm evaluation, and re-implementation of existing methods. Furthermore, we demonstrate that Slax achieves competitive performance compared to other frameworks while offering greater flexibility in implementing complex learning rules.



\section{Related Work}

\begin{table}
    \centering
    \begin{tabular}{ccccccccc}
         & \ssmall BPTT&  \ssmall RTRL&  \ssmall FPTT&  \ssmall OTTT& \ssmall OSTL& \ssmall OTPE& \ssmall STDP& \ssmall ANN2SNN\\
 \ssmall snnTorch& \cmark&   \cmark&  \xmark&  \xmark&  \xmark& \xmark& \cmark& \xmark\\
 \ssmall Spyx& \cmark& \xmark& \xmark& \xmark& \xmark& \xmark& \xmark& \xmark\\
 \ssmall BindsNet& \cmark& \xmark& \xmark& \xmark& \xmark& \xmark& \cmark& \xmark\\
 \ssmall SpikingJelly& \cmark& \xmark& \cmark& \cmark& \xmark& \xmark& \cmark& \cmark\\
 \ssmall Norse& \cmark& \xmark& \xmark& \xmark& \xmark& \xmark& \cmark& \xmark\\
 \ssmall Slax& \cmark& \cmark& \cmark& \cmark& \cmark& \cmark& *\xmark & *\xmark \\
 \ssmall Rockpool& \cmark& \xmark& \xmark& \xmark& \xmark& \xmark& \xmark& \xmark\\
    \end{tabular}
    \caption{A non-exhaustive table of training algorithms provided out-of-the-box by several popular SNN packages. The algorithms with an asterisk will be implemented in the future.}
    \label{tab:algs}
\end{table}


The rich ecosystem of SNN packages caters to a range of needs such as bio-plausibility, deployment on specialized hardware, or rapid prototyping and research. Slax focuses on the latter category, aiming to facilitate research into performant learning algorithms. While, there is some overlap with existing approaches, Slax coexists in the ecosystem, with several popular frameworks listed below:
\begin{itemize}
    \item snnTorch \cite{snnTorch}: One of the most widely used PyTorch-based SNN training packages, known for its extensive collection of neuron models, surrogate derivatives, and  scoring and performance metrics. SnnTorch primarily focuses on BPTT with limited support for alternative training algorithms.
    
    \item SpikingJelly \cite{spikingjelly}: A PyTorch-based SNN package offering high-performance implementations of certain neuron models with custom Cuda and CuPy kernels. This package also offers multiple training algorithms, albeit sometimes with certain model restrictions and limited performance.
    
    \item Spyx \cite{spyx}: A recent JAX-based SNN package prioritizing efficient training and execution using JAX's JIT compilation and data pre-staging. Spyx offers a reduced memory footprint and increased speed for BPTT-based training, with limited support for exploring diverse training algorithms.
    
    \item BindsNet \cite{bindsnet}: This framework emphasizes biological plausibility and supports implementation on the SpiNNaker neuromorphic platform~\cite{spinnaker}. BindsNet offers STDP-based learning rules, including two-factor and three-factor variants, as well as custom implementation. However, its focus is distinct from Slax's emphasis on ease of prototyping performant learning algorithms.
    
    \item Norse \cite{norse}: A PyTorch-based library designed for deep learning with SNNs, offering a diverse collection of biologically plausible neuron models, synapse dynamics, and learning rules. Norse primarily emphasizes using implementing deep learning-based techniques for SNNs. 
    
\end{itemize}

\begin{figure}[h]

\centering
    
\subfloat[]{
\includegraphics[width=0.27\textwidth]{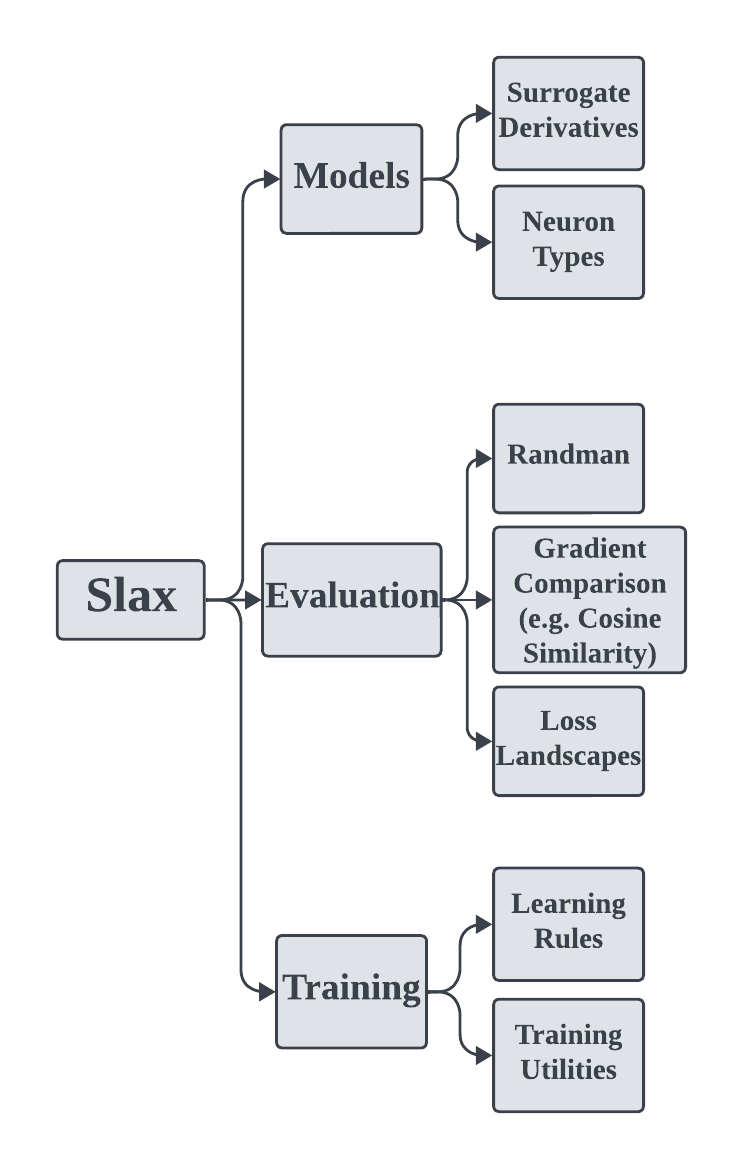}\hfill
}
\subfloat[]{
\includegraphics[width=0.7\textwidth]{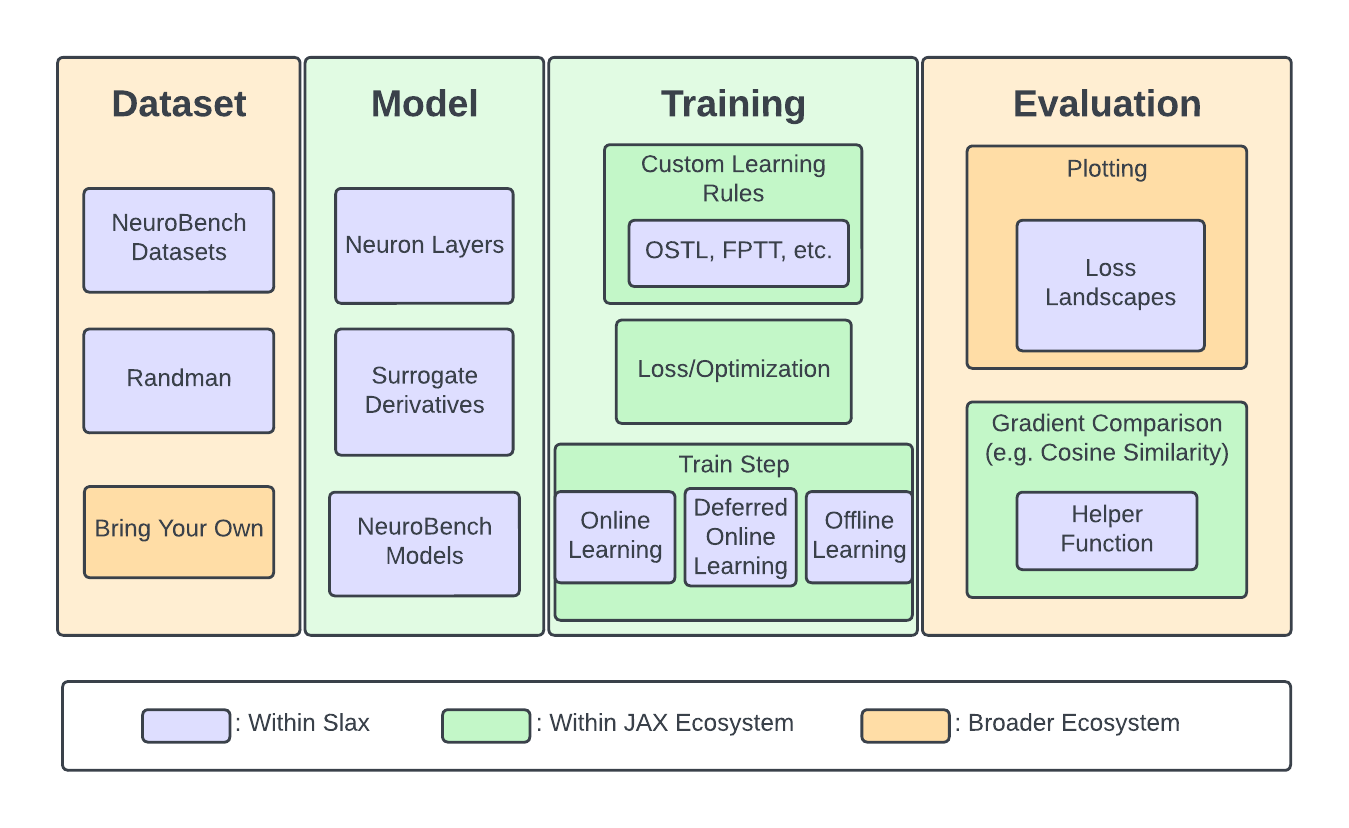}\hfill
}
\caption{Diagram of Slax components. Figure (a) shows the library's structure while figure (b) shows the design and implemented components. While datasets and dataloaders may come from other packages such as tensorflow-datasets~\cite{TFDS}, Slax directly provides NeuroBench datasets and the Randman synthetic dataset. All composable and differentiable functions should use JAX-based libraries or pure Python, but some neuron layers, surrogate derivatives, and NeuroBench models are easily accessible to the user. The same applies to defining a training loop, where Slax provides custom learning rules and easy-to-use training loops for online and offline learning. Evaluating networks is open-ended, but Slax provides utilities for loss landscapes and comparing gradients, which are useful for comparing learning rules.}
\label{fig:slax}

\end{figure}

\section{Design and Usability}

\begin{figure}[ht!]
\centering
\begin{tabular}{c}
\begin{minipage}[c]{.45\linewidth}
\begin{minted}[bgcolor=bg, fontsize=\footnotesize, breaklines, linenos,autogobble,
    framesep=3mm]{python}
class LIF(SNNCell):
  init_tau: float = 2.
  spike_fn: Callable = fast_sigmoid()
  v_threshold: float = 1.0
  v_reset: float = 0.0
  dtype: Any = jnp.float32
  carry_init: nn.initializers.Initializer = nn.initializers.zeros_init()

  @nn.compact
  def __call__(self,carry,x=None):
    if x == None:
      x = carry
      Vmem = self.variable('carry', 'Vmem', self.carry_init, jax.random.PRNGKey(0), x.shape, self.dtype )
      vmem = Vmem.value
      hidden_carry = True
    else:
      vmem = carry['Vmem']
      hidden_carry = False
\end{minted}
\end{minipage}\hspace{0.1\linewidth} 
\begin{minipage}[c]{.45\linewidth}
\begin{minted}[bgcolor=bg, fontsize=\footnotesize, breaklines, linenos, firstnumber=19,autogobble,
    framesep=3mm]{python}
    vmem = nn.sigmoid(tau) * vmem + x
    spikes = self.spike_fn(vmem - self.v_threshold)
    vmem -= spikes * self.v_threshold

    if self.is_initializing() and x==None:
      vmem.value = self.carry_init( jax.random.PRNGKey(0), x.shape, self.dtype)

    if hidden_carry:
      Vmem.value = vmem
      return spikes
    else:
      carry['Vmem'] = vmem
      return carry, spikes
\end{minted}
\end{minipage} 
\end{tabular}
\cprotect\caption{The code above defines a Slax LIF neuron in the style neurons are written in the library. All neurons are implemented with extra logic for handling an explicit state to match Flax RNNs while inferring the neuron state's shape.}
\label{fig:slax_snn}
\end{figure}

Slax is designed to facilitate the exploration and implementation of diverse training algorithms for SNNs, with a strong emphasis on flexibility and efficiency. Figure~\ref{fig:slax_snn} illustrates an example of the Slax style for implementing a Leaky Integrate-and-Fire neuron. Slax modules only deviate from Flax where these deviations might enhance usability for SNNs. Despite deviations, they maintain bidirectional compatibility with Flax, allowing for seamless integration within the broader JAX ecosystem. Slax offers versatile tools that can be readily applied to Flax recurrent neural networks (RNNs) with minimal effort, and conversely, Slax modules can function as Flax RNNs without modifications. This allows Slax generated SNNs to be interoperable with existing JAX and Flax workflows. Additionally, Slax simplifies the creation of SNNs with custom learning rules through a set of composable functions. Primary amongst these, the \texttt{connect} function enables the flexible construction of recurrent SNN architectures linking Flax and Slax modules.

Slax offers flexibility in implementing new neuron models. While the simplest approach involves extending the Flax \texttt{RNNCellBase} class, see Fig.~\ref{fig:flax_snn}, the neuron models provided within the package include additional logic and functions to enable automatic shape inference for spiking layers. Using the Slax conventions as shown enhances usability for the typical user, reducing boilerplate code.

\begin{figure}[h]
\centering
\begin{tabular}{c}
\begin{minipage}[c]{.45\linewidth}
\begin{minted}[bgcolor=bg, fontsize=\footnotesize, breaklines, linenos, autogobble,
    framesep=3mm]{python}
class FlaxLIF(RNNCellBase):
  features: int
  init_tau: float = 2.
  v_thresh: float = 1.0
  spike_fn: Callable = fast_sigmoid()
  dtype: Any = jnp.float32
  carry_init: Any = nn.initializers.zeros_init()

  @nn.compact
  def __call__(self,carry):
    vmem = carry['Vmem']
    vmem = nn.sigmoid(tau) * vmem + x
    spikes = self.spike_fn( vmem - self.v_thresh )
    vmem -= spikes * self.v_threshold
    carry['Vmem'] = vmem
    return carry, vmem
\end{minted}
\end{minipage}\hspace{.09\linewidth}
\begin{minipage}[c]{.45\linewidth}
\begin{minted}[bgcolor=bg, fontsize=\footnotesize, breaklines, linenos, firstnumber=17, autogobble,
    framesep=3mm]{python}
  @nowrap
  def initialize_carry( self, rng: PRNGKey, input_shape ):
    batch_dims = input_shape[:-1]
    mem_shape = batch_dims + (self.features,)
    carry = {'Vmem': self.carry_init(rng, mem_shape, self.dtype)}
    return carry
    
LIF = flax_wrapper( FlaxLIF( features = 1000) )
\end{minted}
\end{minipage}

\end{tabular}
\cprotect\caption{This code achieves the same result as Fig. \ref{fig:slax_snn} but instead follows the typical Flax RNN conventions. A wrapping function is then applied to the defined model to produce a Slax-compatible neuron. While this requires less code, the layer features must now be specified when defining the layer.}
\label{fig:flax_snn}
\end{figure}

\begin{figure}[htb]
\centering
\begin{tabular}{c}
\begin{minipage}[c]{.65\linewidth}
\centering
\includegraphics[width=0.9\textwidth]{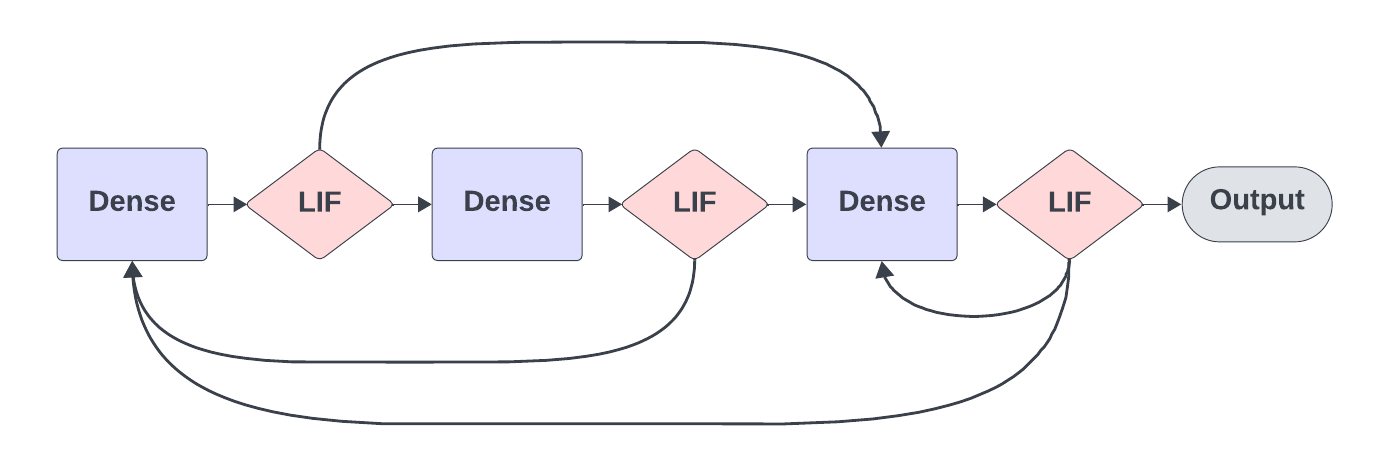}
\end{minipage}
\begin{minipage}[c]{.3\linewidth}
\begin{minted}[bgcolor=bg, fontsize=\footnotesize, linenos,autogobble,
    framesep=3mm]{python}
sl.connect([nn.Dense(100),
            sl.LIF(),
            nn.Dense(100),
            sl.LIF(),
            nn.Dense(100),
            sl.LIF()],
            cat={'0':[3,5],
            '4':[1,5]]})
\end{minted}
\end{minipage}
\end{tabular}
\cprotect\caption{A diagram of an SNN with complex recurrent connections and its accompanying code. The \texttt{connect} function allows the user to specify recurrent and skip connections.}
\label{fig:connect}
\end{figure}

In the same vein as other SNN frameworks, Slax provides implementations of various neuron models and surrogate gradient functions. Surrogate gradients are key for SNN training with automatic differentiation (AD) due to the non-differentiable nature of the Heaviside step function. Slax includes commonly used surrogate gradients, and its composability eases the development of custom functions that can leverage performant AD from JAX and Flax. In addition to building with Flax's efficient implementations for linear connection functions (the model weights), Slax provides tools for spiking neuron layers. Figures ~\ref{fig:connect} and ~\ref{fig:snnModel} demonstrate how Flax and Slax modules can be combined to define trainable SNN models.

\paragraph{Recurrence:} Feed-forward SNNs exhibit recurrence through the dynamics of the membrane potential, even without explicit trainable recurrent parameters. Although feedforward SNNs are effective for many tasks, introducing recurrence via self-loops or feedback connections can be particularly effective in vision and audio domains~\cite{SHD,implicit-snn}. Slax neurons inherently capture recurrent effects within the neuron state. However, due to the separation of linear connection functions and neuron layers, additional recurrent connections require explicit definition. The \texttt{connect} function, as shown in Figure~\ref{fig:connect}, facilitates the easy implementation of such recurrent loops within SNN architectures. 


\begin{figure}[h]
\begin{tabular}{c}
\begin{minipage}[c]{.48\linewidth}
\begin{minted}[bgcolor=bg, fontsize=\footnotesize, breaklines, linenos,autogobble,
    framesep=3mm]{python}
import jax.numpy as jnp
import flax.linen as nn
import slax as sl

class snnModel(nn.Module):
    @nn.compact
    def __call__(self,x):
        x = nn.Dense(20)(x)
        x = nn.RNN(sl.LIF( tau=0.7, surrogate=sl.atan()))(x)
        x = nn.Dense(10)(x)
        x = nn.RNN(sl.LIF( tau=jnp.arange(1,0.1), trainable_tau=True))(x)
        return x
\end{minted}
\end{minipage}
\begin{minipage}[c]{.48\linewidth}
\vspace{.2\linewidth}
\begin{minted}[bgcolor=bg, fontsize=\footnotesize, breaklines, linenos, autogobble,
    framesep=3mm]{python}
class snnModel(nn.Module):
    @nn.compact
    def __call__(self,x):
        x = nn.Dense(20)(x)
        x = nn.RNN(FlaxLIF(tau=0.7, surrogate=sl.atan()))(x)
        x = nn.Dense(10)(x)
        x = nn.RNN(FlaxLIF( jnp.arange(1,0.1), trainable_tau=True))(x)
        return x
\end{minted}
\end{minipage} \\
\end{tabular}
\cprotect\caption{The code above defines a Flax SNN model. The left code block uses Slax LIF neurons while the right code block uses a Flax \verb|RNNCellBase| from Fig.~\ref{fig:flax_snn}. Both are functionally equivalent and are fully compatible with Flax's API for RNNs, such as \verb|flax.linen.RNN|, which loops the layer and its state through the time dimension of the input.}
\label{fig:snnModel}
\end{figure}

\paragraph{Parity with Flax:} Ensuring compatibility and parity with Flax is a core design principle of Slax. Neuron models in Slax are designed to function seamlessly as Flax RNN cells, allowing for easy integration within Flax models and training routines. Slax handles neuron state variables as part of the Flax variable collection while also offering optional support for explicit state input and output. This design choice provides flexibility for users, allowing them to either rely on Slax's automatic state management or directly use Flax RNN functionalities as needed.

Additionally, Slax uniquely defines surrogate derivatives to support both forward- and reverse-mode AD. Figure~\ref{fig:define_surrogate} illustrates an example of defining a forward-mode compatible surrogate derivative in Slax. This enables mixed-mode gradient calculations and efficient Hessian computations, expanding the possibilities for advanced optimization techniques. By supporting both vector-Jacobian and Jacobian-vector products, custom derivatives in Slax achieve a level of flexibility comparable to any other differentiable function within the JAX ecosystem.


\begin{figure}[h]
\begin{tabular}{c}
\begin{minipage}[t]{.48\linewidth}
\begin{minted}[bgcolor=bg, fontsize=\footnotesize, breaklines, linenos,autogobble,
    framesep=3mm]{python}
def fast_sigmoid(slope=25):
    @jax.custom_jvp
    def fs(x):
      return jnp.array(x >= 0.0,
                  dtype=x.dtype)
\end{minted}
\end{minipage}
\begin{minipage}[t]{.48\linewidth}
\begin{minted}[bgcolor=bg, fontsize=\footnotesize, breaklines, linenos, firstnumber=6, autogobble,
    framesep=3mm]{python}
    @fs.defjvp
    def fs_fwd(primal, tangent):
        x, = primal
        t, = tangent
        alpha = slope
        scale = 1/(alpha*jnp.abs(x)+1.)**2
        return (fs(x),scale*t)
\end{minted}
\end{minipage} \\
\end{tabular}
\cprotect\caption{The code above defines the fast sigmoid function surrogate from SuperSpike \cite{superspike}. The surrogate derivative is written as a Jacobian-vector product, which JAX can use for both forward and reverse-mode AD.}
\label{fig:define_surrogate}
\end{figure}

\subsection{Learning Rules}

Offline training using BPTT with surrogate gradients and feed-forward architectures is the most commonly employed usecase for SNN research. Slax aims to maintain performance and ease of use for such users while supporting a wider range of learning algorithms. Building on existing conventions for implementing Flax and Optax, Slax also simplifies more unconventional exploratory approaches. Figure~\ref{fig:bptt_model} illustrates a single BPTT update step for a model in Slax.

\begin{figure}[hbp]
    \centering
\begin{minted}[bgcolor=bg, fontsize=\footnotesize, breaklines, linenos,  autogobble,
    framesep=3mm]{python}
batch = sl.randman(manifold_PRNG,random_PRNG) # Generate a batch for randman
key = jax.random.PRNGKey(0) # initialize a pseudo-random key
snn = sl.connect([nn.Dense(50), # create an SNN with two dense layers
                  sl.LIF(),
                  nn.Dense(10),
                  sl.LIF()])
params = snn.init(key, batch[0]) # initialize model parameters with an input array
optimizer = optax.adamax(0.001) # set the optimizer
opt_state = optimizer.init(params['params']) # initialize the optimizer state
train_step = sl.train_offline(snn,
                              optax.softmax_cross_entropy,
                              optimizer) # Initialize an offline training module
updated_params,updated_opt_state,spikes,loss = train_step(params,
                                                          opt_state,
                                                          batch) # Applies forward+backward pass
\end{minted}
    \caption{Full code for generating a batch of data sampled from Randman, initializing an SNN, and performing a single forward+backward step.}
    \label{fig:bptt_model}
\end{figure}

\begin{figure}[h]
\centering
\begin{minipage}[]{\linewidth}
\begin{minted}[bgcolor=bg, fontsize=\footnotesize, breaklines, linenos, autogobble,
    framesep=3mm]{python}
snn = sl.connect([nn.Dense(50),
                  sl.LIF(),
                  nn.Dense(10),
                  sl.LIF()],cat={0:[1]}) # concatenates output of layer 1 to the input of layer 0
train_step = sl.FPTT(snn,
                     optax.softmax_cross_entropy,
                     optimizer) # applies FPTT training algorithm to the model
params,opt_state,spikes,loss = train_step(params,opt_state,input_array)
\end{minted}
\end{minipage}
\begin{minipage}[]{\linewidth}
\begin{minted}[bgcolor=bg, fontsize=\footnotesize, breaklines, linenos, firstnumber=9, autogobble,
    framesep=3mm]{python}
snn = sl.connect([sl.DenseOSTL([nn.Dense(50), # Applies OSTL to list of layers
                  sl.LIF()]),
                  sl.DenseOSTL([nn.Dense(10),
                  sl.LIF()])])
train_step = sl.train_online(snn,optax.softmax_cross_entropy,optimizer)
params,opt_state,spikes,loss = train_step(params,opt_state,input_array)
\end{minted}
\end{minipage} \\
\cprotect\caption{Slax's \verb|connect| function defining SNN models. Learning rules, such as OSTL can be applied within or on top of networks created by the \verb|connect| function. In the top code block, the function's optional \verb|cat| argument allows specifying recurrent or skip connections to any other layer by concatenating (technically initializing a new layer) the output of a layer to the input of another.}
\label{fig:learning_rules}
\end{figure}

As we previously discussed, Slax can significantly reduce code complexity for training SNNs. As shown in Figure~\ref{fig:learning_rules}, Slax can reduce the amount of code required to implement different training algorithms or incorporate recurrent models with only minimal code modifications. Slax provides a range of online learning algorithms and utilities for their efficient implementation. Contemporary research on SNN training has complex requirements across dimensions like bio-plausibility, efficiency, and online learning capabilities. Slax currently supports a range of these algorithms including Online Spatio-Temporal Learning (OSTL) \cite{ostl}, Online Training Through Time \cite{ottt}, OTPE \cite{otpe}, Real-Time Recurrent Learning (RTRL) \cite{rtrl}, and Forward Propagation Through Time (FPTT) \cite{fptt}.

Slax provides the \texttt{train\_online} utility function to facilitate online learning, where network parameters are updated at each time-step within the training loop. FPTT, unlike other online algorithms, acts as a regularizer rather than a direct gradient calculation method. To accommodate this, Slax offers the \texttt{slax.FPTT} function, which can be used in conjunction with other online algorithms or as a replacement for \texttt{train\_online}.


The code example provided in Figure~\ref{fig:learning_rules} demonstrates the implementation of OSTL for dense layers. Slax includes an optimization for dense layers that leverages the sparsity patterns inherent to these layers, avoiding the overhead associated with general sparse operations. Layer-local learning rules, such as OSTL and OTTT, operate on a single Flax module of a chain of modules, returning a modified module that is mutated by the learning rule. Unlike layer-local learning rules, which must be repeatedly applied in deep networks, Slax's implementation of RTRL can be applied to all the layers in the model.

\begin{figure}[H]

\centering
\subfloat[]{
\includegraphics[width=.5\textwidth]{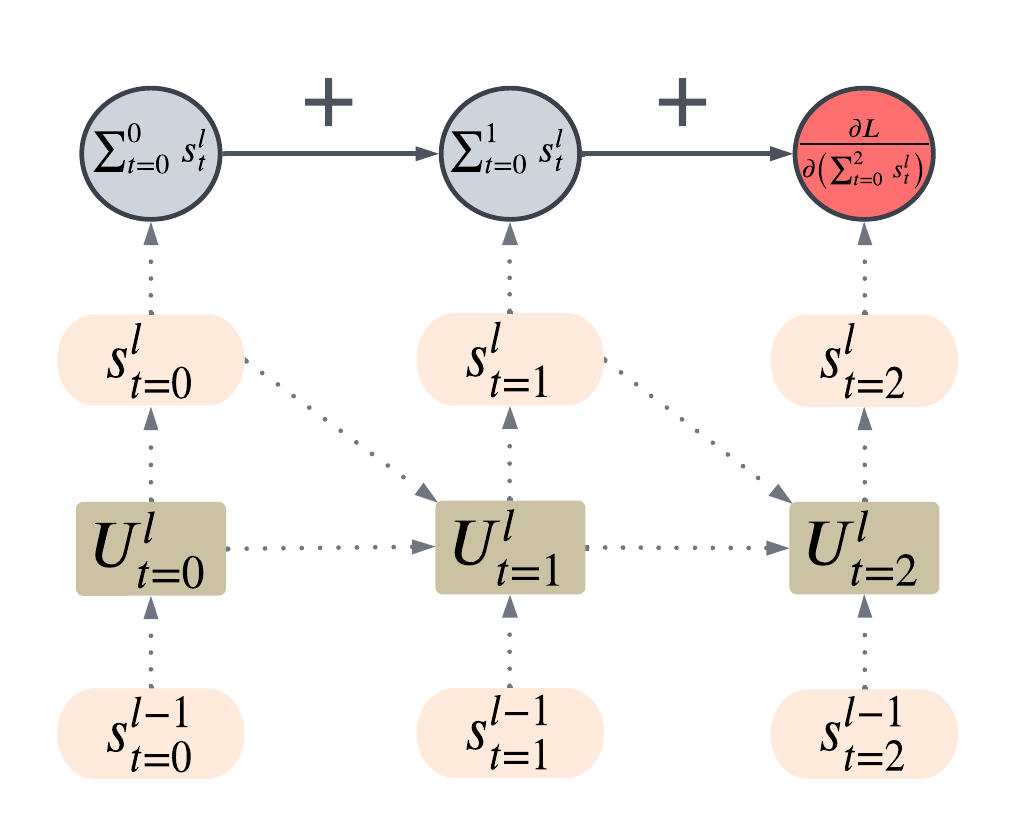}\hfill}
\subfloat[]{
\includegraphics[width=.5\textwidth]{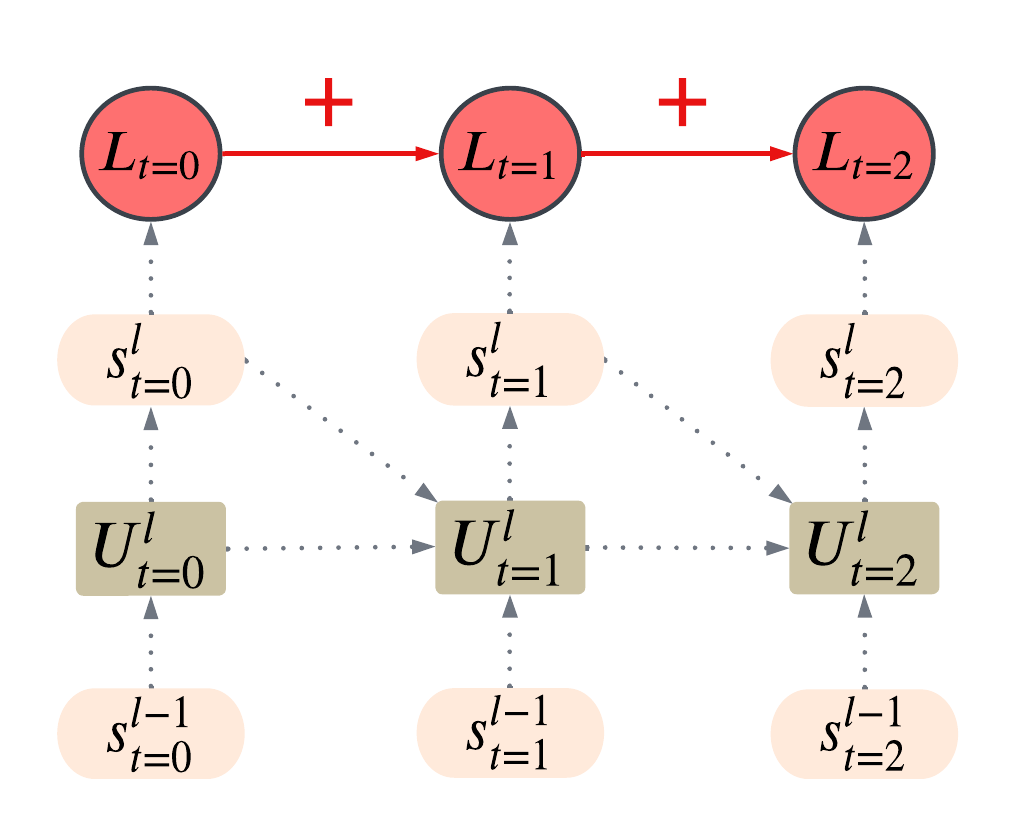}}
\caption{(a) Calculating loss as normally done when training offline. The output spikes are accumulated across time-steps, and then loss, such as softmax cross entropy, is applied to the accumulated spikes. (b) Calculating loss at each time-step. Instead of accumulating spikes, the loss function is applied to the output at each time-step, and the loss is accumulated through time instead.}
\label{fig:scoring}

\end{figure}

Slax enables the use of online learning rules in the offline learning setting through the \texttt{train\_online\_deferred} function which delays the parameter updates until the final timestep. Figure~\ref{fig:scoring} illustrates a key distinction between online and offline training: online learning calculates loss at each time step, while offline learning (typically with BPTT) computes loss only at the end based on accumulated spikes. This difference prevents easy correspondence between BPTT and online learning rules. Slax provides an Optax \cite{optax} loss function \texttt{train\_offline} to calculate the equivalent loss.

Slax simplifies the process of creating new learning rules by providing helper functions to enable rapid algorithm prototyping. While some of these functions are under development, Slax aims to offer tools that handle batch dimensions, manage multiple state variables, reduce boilerplate code, and assist in navigating the complexities of defining custom gradients within JAX.  We have developed Slax to enable JAXs easy vectorization across the batch dimension, such that ideally, custom gradient calculations can be implemented without explicit batch dimensions and with a single state variable. This approach streamlines matrix operations and eases the translation from algorithm pseudo-code to functional JAX/Flax/Slax code.


\subsection{Visualization and Metrics}

\begin{figure}[H]

\centering
\subfloat[]{
\includegraphics[width=0.63\textwidth]{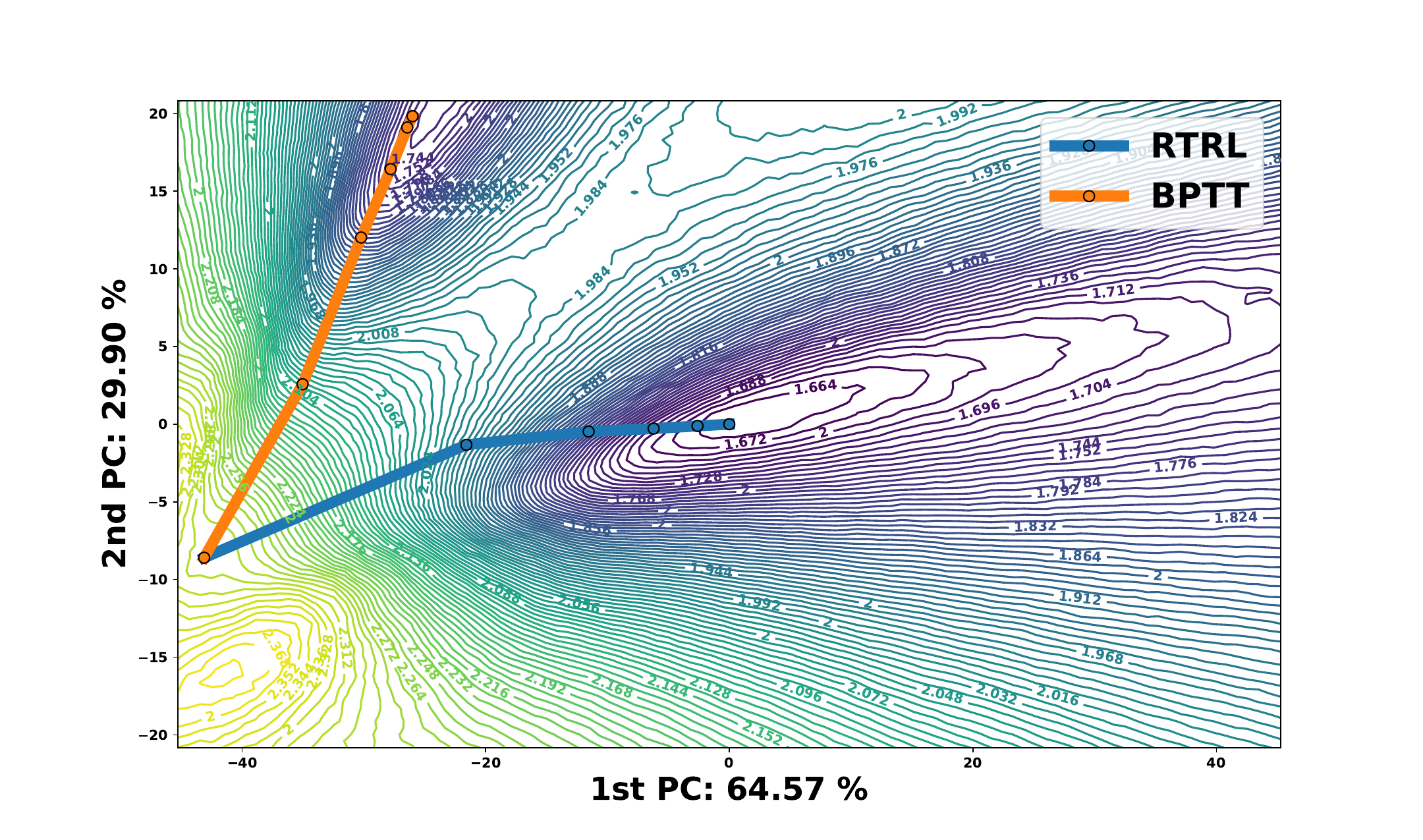}\hfill}
\subfloat[]{
\includegraphics[width=.35\textwidth]{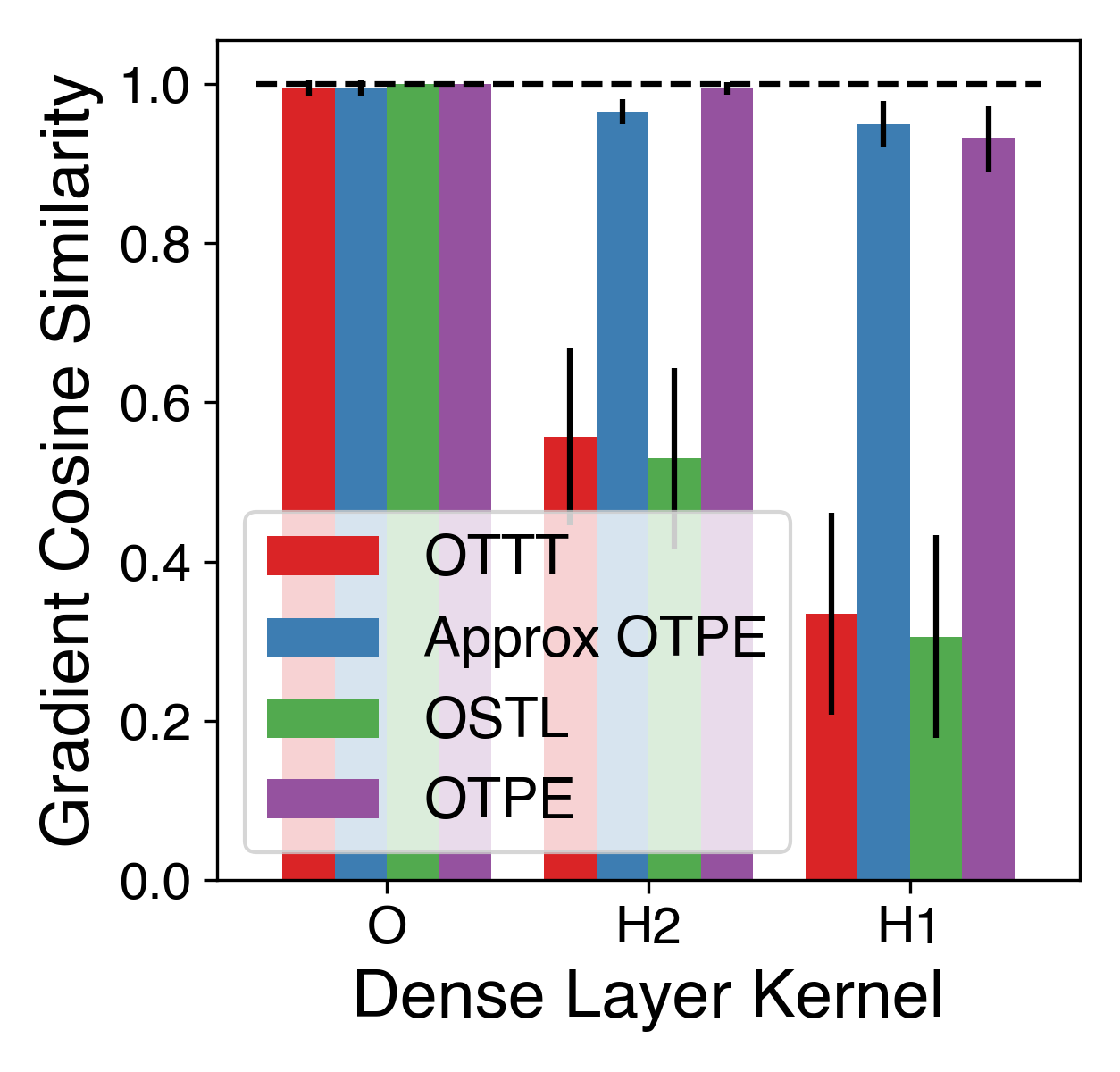}}
\caption{(a) sample loss landscape plot comparing algorithms on rate-encoded randman. The RTRL implementation is updating online while BPTT is offline. (b) Plot of cosine similarity between BPTT and other algorithms for rate-encoded randman. This is one application of comparing gradients with the \texttt{compare\_grads} function.}
\label{fig:training_visualization}

\end{figure}

In addition to providing support for executing online learning, we also provide utilities for evaluating these algorithms. We provide a utility for generating the loss landscape for a model throughout training, as described in \citet{loss_landscapes}. Our function supports plotting one or more training trajectories, as seen in Fig. \ref{fig:training_visualization} (a). We also provide a function for comparing gradients between two methods updating the same model. We provide cosine similarity measures as a default.

We also include a dataset to evaluate how well an algorithm learns rate vs time encoded information. Randman \cite{randman} is a synthetic dataset, aimed at studying SNN capabilities in learning spike timing patterns. Spike-times are generated on a random smooth manifold, projected onto a high-dimensional hypercube. These are then sampled to generate a spike-train, which the SNN learns to classify. The dimensionality and smoothness of the manifold are adjustable to change the task difficulty. Neurons, firing once per trial, only contain temporal information, a format we term T-Randman. This is also the original design. We modify Randman to test rate-based learning through a format we term R-Randman. Here, spike-rates are determined by manifold values and are generated to be temporally uncorrelated through random shuffling.

While Randman allows for a direct comparison of learning rate and time encoded information, it does not speak to performance on real tasks. For this reason, we add compatibility with NeuroBench and its test harness. Training with NeuroBench has effectively the same implementation as using a Torch dataloader, but we also export a reimplementation of NeuroBench's testing harness that is compatible with Slax SNNs. The harness can provide many metrics beyond accuracy, such as activation sparsity and the number of neuron updates. This provides a standard method of comparison while providing efficiency metrics, which can be of interest for research on learning rules \cite{ottt,implicit-snn}.

\section{Performance}

\begin{figure}[H]

\centering


\includegraphics[width=.51\textwidth]{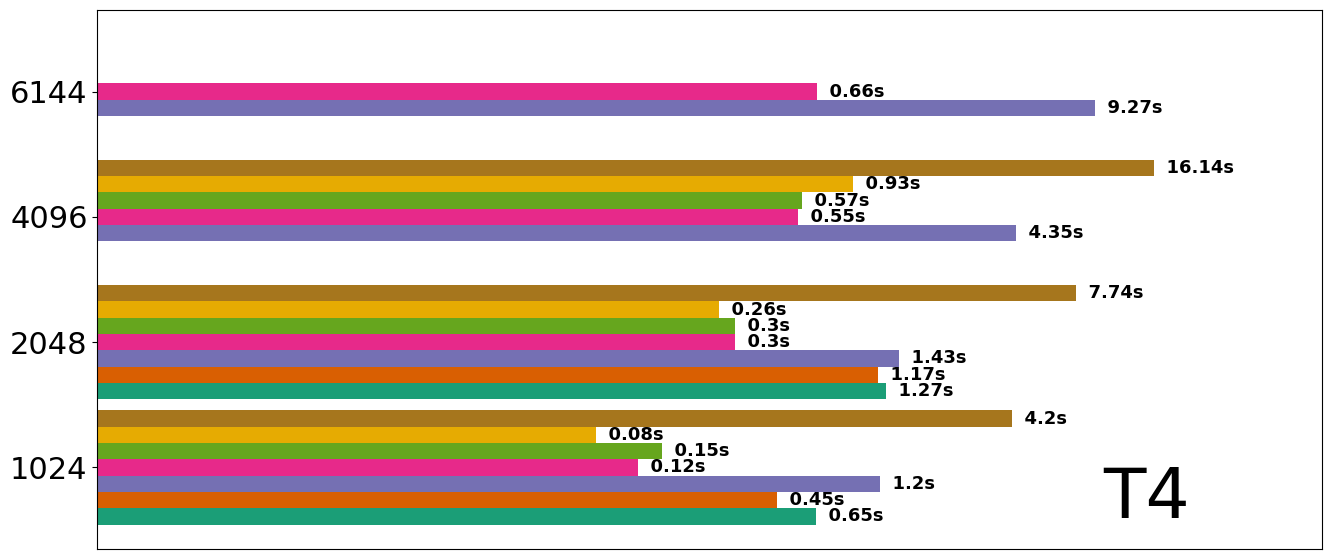}
\includegraphics[width=.49\textwidth]{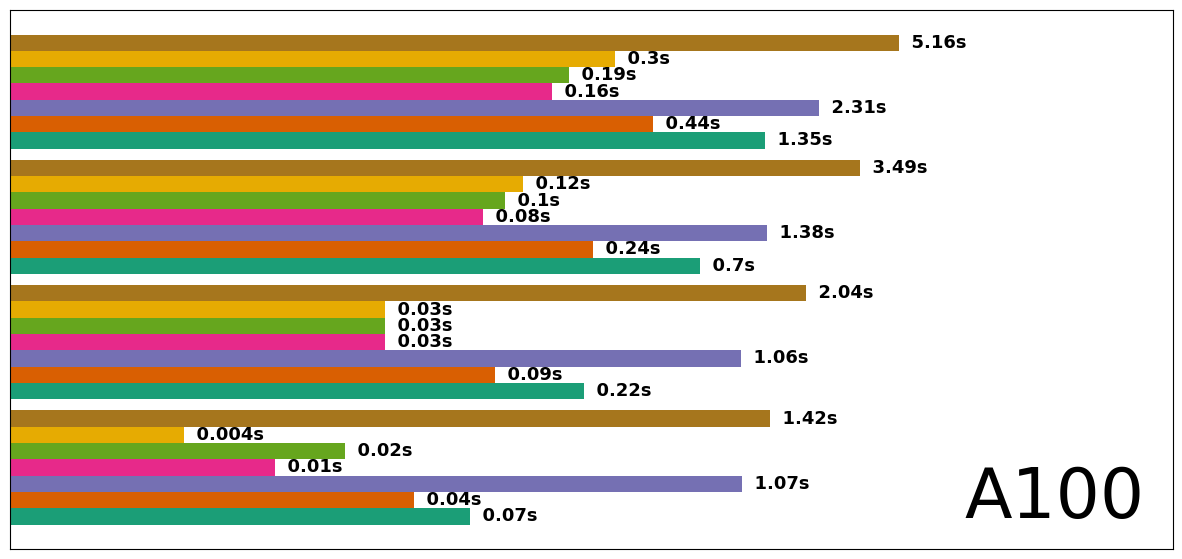}\hfill
\includegraphics[width=.51\textwidth]{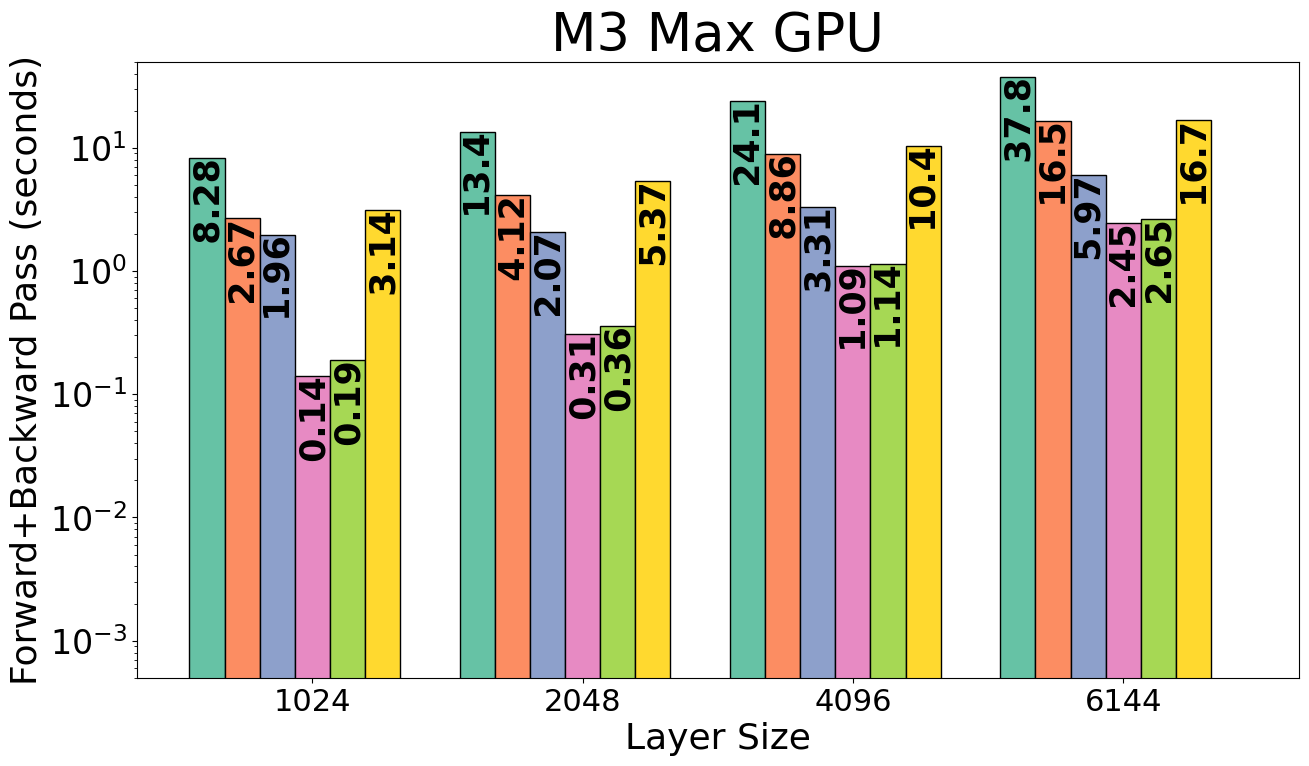}
\includegraphics[width=.49\textwidth]{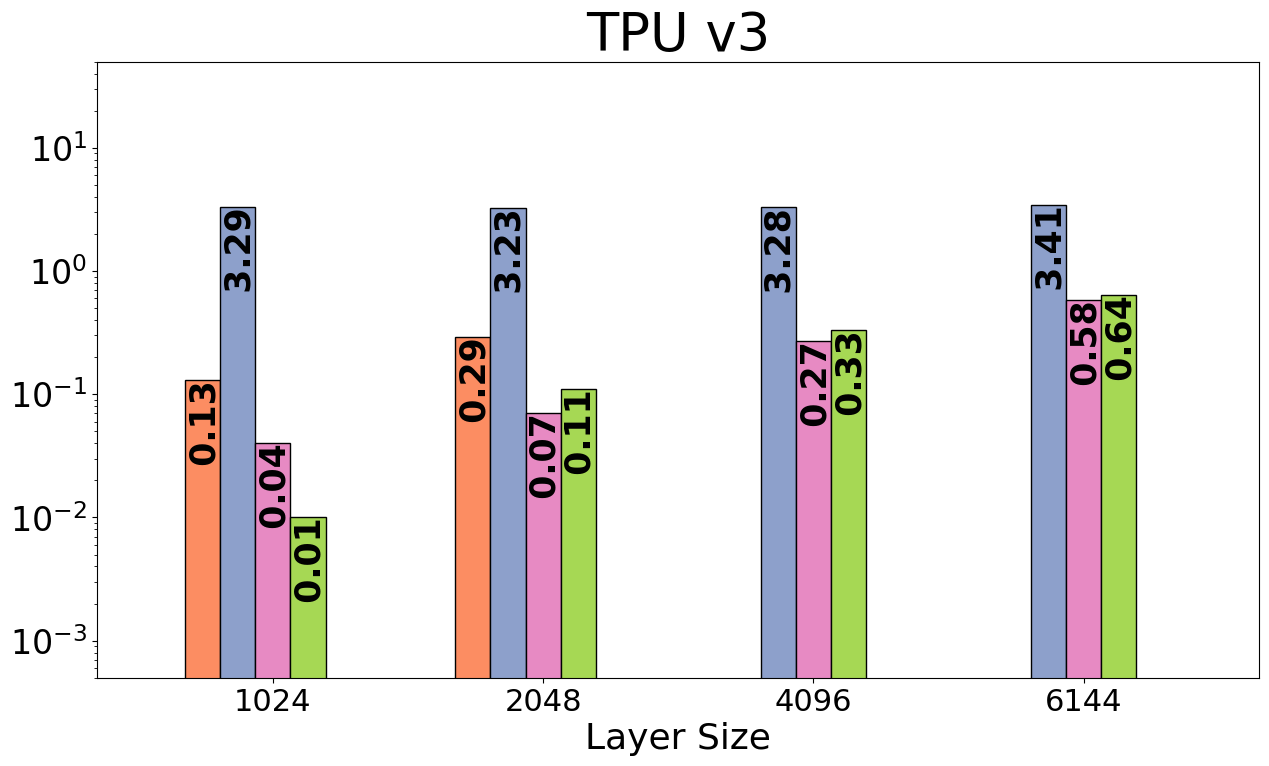}

\includegraphics[width=1.\textwidth]{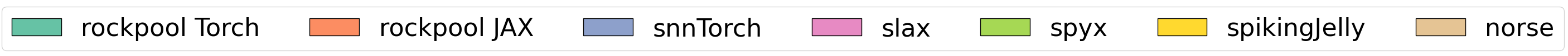}

\caption{Speed of the combined forward and backward pass of multiple SNN libraries in seconds. The reported is the average of 20 runs using a three Dense layer network with a sequence length of 500 and batch size of 64.}
\label{fig:performance}

\end{figure}

Although we prioritize online training capability over training speed and resource consumption, Slax is competitive with Spyx and can outperform PyTorch-based frameworks, as seen in Fig. \ref{fig:performance}. Even using custom CUDA code, SpikingJelly does not achieve the same level of scalability as Slax or Spyx. These performance results, however, are only representative of basic implementations for each package and may not reflect their best performance capabilities. Importantly, CUDA implementations of neurons, such as in EXODUS \cite{exodus} and the SpikingJelly CuPy LIF, only support feed-forward SNNs. JAX-based SNNs can achieve the speed of custom CUDA code while having the flexibility to define custom recurrent connections.

\section{Discussion}

While Slax already implements many learning rules and achieves competitive efficiency, the package is far from complete. In addition to expanding our current selection of neuron types, training algorithms, and surrogate derivatives, we also intend to add support for more complex functionality. 
\begin{itemize}
    \item Adjustable Randman. Randman can currently be rate or time encoded, but only being one or the other does not reflect real datasets, which normally contain information encoded both ways. To rectify this, we will implement a version of Randman that allows a specified percentage of information to be rate-encoded.
    \item Sparsity. To take full advantage of SNN sparsity, especially on long time sequences, we will be adding layer-wise support for the JAX experimental sparsity module.
    \item Alternatives to surrogate methods. There are cases where gradients can be calculated without the use of surrogate derivatives, such as EventProp~\cite{eventprop}, implicit differentiation in feedback SNNs \cite{implicit-snn}, and scoring pseudo-spikes in QIFs \cite{pseudo-spike}.
    \item Convolutions for custom learning rules. While some custom learning rules are compatible with any linear connection by default, learning rules such as OSTL do not have an efficient implementation for convolutional layers.
\end{itemize}

Slax, an SNN framework designed to research training SNNs, brings performance comparable to or better than other SNN frameworks while also enabling higher flexibility. Support for forward-mode differentiation, online learning, and methods for evaluating training makes Slax stand out from other packages when it comes to training SNNs. Despite Slax already being more flexible than most, a substantial number of items remain to be implemented. Aside from expanding our neuron and surrogate types, we are also developing a flexible Randman dataset, which can smoothly transition between rate and time encoding, sparse computation to reduce the cost of long time sequences, and more training algorithms (e.g. predictive coding and STDP). We welcome any feedback and contributions to our package.

\newpage

\bibliography{refs}

\end{document}